\documentclass[10pt,twocolumn,letterpaper]{article}
\usepackage{cvpr}
\usepackage{times}
\usepackage{graphicx}
\usepackage{amsmath}
\usepackage{amssymb}
\usepackage{cases}
\usepackage{booktabs}
\usepackage{setspace}
\graphicspath{{figures/}}
\usepackage[sort,numbers]{natbib}
\usepackage{subeqnarray}
\usepackage{multirow}
\usepackage{tabularx}
\usepackage{epstopdf}
\usepackage{subfigure}

\usepackage{array}
\newcolumntype{I}{!{\vrule width 3pt}}
\newlength\savedwidth

\newlength\savewidth
\newcommand\shline{\noalign{\global\savewidth\arrayrulewidth
                            \global\arrayrulewidth 0.8pt}%
                   \hline
                   \noalign{\global\arrayrulewidth\savewidth}}

\newcolumntype{C}{>{\centering\arraybackslash}X}

\makeatother

\usepackage[pagebackref=true,breaklinks=true,letterpaper=true,colorlinks,bookmarks=false]{hyperref}

\cvprfinalcopy 

\def\cvprPaperID{} 

\begin{document}

\title{Deep Plug-and-Play Super-Resolution for Arbitrary Blur Kernels}

\author{Kai Zhang$^{1,2}$,  Wangmeng Zuo$^{1,3,}$\thanks{Corresponding author.} , Lei Zhang$^{2,4}$\\
$^1$School of Computer Science and Technology, Harbin Institute of Technology, Harbin, China\\
$^2$Dept. of Computing, The Hong Kong Polytechnic University, Hong Kong, China\\
$^3$Peng Cheng Laboratory, Shenzhen, China\\
$^4$DAMO Academy, Alibaba Group\\
{\tt\small cskaizhang@gmail.com, wmzuo@hit.edu.cn, cslzhang@comp.polyu.edu.hk}\\
\small \url{https://github.com/cszn/DPSR}
}

\maketitle
\thispagestyle{empty}

\begin{abstract}

While deep neural networks (DNN) based single image super-resolution (SISR) methods are rapidly gaining popularity, they are mainly designed for the widely-used bicubic degradation, and there still remains the fundamental challenge for them to super-resolve low-resolution (LR) image with arbitrary blur kernels.
In the meanwhile, plug-and-play image restoration has been recognized with high flexibility due to its modular structure for easy plug-in of denoiser priors.
In this paper, we propose a principled formulation and framework by extending bicubic degradation based deep SISR with the help of plug-and-play framework
to handle LR images with arbitrary blur kernels.
Specifically, we design a new SISR degradation model so as to take advantage of existing blind deblurring methods for blur kernel estimation.
To optimize the new degradation induced energy function, we then derive a plug-and-play algorithm via variable splitting technique, which allows us to plug any super-resolver prior rather than the denoiser prior as a modular part.
Quantitative and qualitative evaluations on synthetic and real LR images demonstrate that the proposed deep plug-and-play super-resolution framework is flexible and effective to deal with blurry LR images.
\end{abstract}

\section{Introduction}
Single image super-resolution (SISR), with the goal of estimating the clean high-resolution (HR) counterpart $\mathbf{x}$ of a given low-resolution (LR) image $\mathbf{y}$, is a classical problem with highly academic and practical values~\cite{baker2002limits}.
Basically, the relationship between LR and HR images is characterized by the degradation model which defines how the LR image is degraded from an HR image.
Empirical and theoretical studies have proven that an accurate degradation model is crucial for the success of SISR~\cite{yang2014single,efrat2013accurate}.
Thus, it is important to first review the degradation model for SISR.

In fact, most existing SISR methods are designed under the assumption of certain degradation model.
There are two widely-used degradation models.
The first one, which is acknowledged as a general degradation model for SISR, is given by
\begin{equation}\label{degradation:eq1}
  \mathbf{y} = (\mathbf{x} \otimes \mathbf{k})\downarrow_s + ~\mathbf{n},
\end{equation}
where $\mathbf{x}\otimes\mathbf{k}$ represents the convolution between blur kernel $\mathbf{k}$ and HR image $\mathbf{x}$,
$\downarrow_s$ is a subsequent downsampling operation with scale factor $s$, and $\mathbf{n}$ is additive white Gaussian noise (AWGN) with noise level $\sigma$.
Such degradation model has been extensively studied on developing model-based optimization methods~\cite{dong2013nonlocally,chan2016plug}.
However, these methods mostly assume a priori known blur kernel which in practice is difficult to estimate.
Although several works such as~\cite{sroubek2008simultaneous,michaeli2013nonparametric} focus on estimating blur kernel, their source codes are unfortunately not publicly available.

The second and perhaps the most widely-used one which we refer to as bicubic degradation is formalized as
\begin{equation}\label{degradation:eq2}
  \mathbf{y} = \mathbf{x}\downarrow_s.
\end{equation}
Here, unless otherwise specified, $\downarrow_s$ instead represents the bicubic downsampler (Matlab default function \texttt{imresize}) with scale factor $s$.
Due to its simplicity, bicubic degradation model has become the benchmark setting to evaluate SISR methods~\cite{kim2015accurate,lim2017enhanced}.
In particular, it greatly facilities the development of powerful deep neural networks (DNN) for SISR~\cite{agustsson2017ntire}.
However, such a simple degradation inevitably gives rise to poor results in many practical scenarios~\cite{efrat2013accurate,zhang2018learning}.
Nevertheless, little work has been done on extending to more realistic degradation models.

Given the above considerations, it is very necessary to address the following two issues:
1) designing an alternative degradation model, and
2) extending existing DNN-based methods for bicubic degradation to the new degradation model so as to exploit the power of DNN.
To this end, we first propose a simple yet effective degradation model which assumes the LR image is a bicubicly downsampled, blurred and noisy version of an HR image.
Compared to the general degradation model given by Eqn.~\eqref{degradation:eq1}, the proposed one offers two favorable merits.
First, it generalizes the well-studied bicubic degradation model given by Eqn.~\eqref{degradation:eq2}.
Second, it enables us to adopt the available blind deblurring methods to estimate the blur kernel from the given LR image.
In order to extend DNN-based SISR methods to the new degradation model, we propose a well-principled deep plug-and-play super-resolution (DPSR) framework
which integrates DNN-based super-resolver into a variable splitting based iterative optimization scheme.
It turns out that the distortion of blur can be effectively handled in Fourier domain.
For this reason, it is possible to deal with arbitrary blur kernels, which is one of the main goals in this paper.
Moreover, different from existing plug-and-play framework~\cite{chan2016plug} which generally plugs the off-the-shelf Gaussian denoiser as a modular part,
the proposed instead achieves the plug-in step by applying any existing DNN-based super-resolvers with only a small modification.

By far, it is worth emphasizing that we mainly focus on non-blind SISR for arbitrary uniform blur kernels rather than blind SISR for arbitrary non-uniform blur kernels.
On the one hand, non-blind SISR is important to blind SISR which generally involves alternatively updating the blur kernel and applying non-blind SISR to update the super-resolved image.
While some recent works attempt to train DNN to directly estimate the clean image for blind deblurring, their practicability needs further evaluations.
On the other hand, although the non-uniform blur kernel tends to be a more realistic assumption, it is too complex and still remains to be a difficult problem for image deblurring~\cite{kruse2017learning}.
In fact, the arbitrary uniform blur kernel assumption is already a much better choice than the simple bicubic kernel for practical applications.
Briefly speaking, our work makes a valuable intermediate step from existing bicubic degradation based SISR to ultimate blind SISR.

The contribution of this work is summarized as follows:
\begin{itemize}
  \vspace{-0.15cm}
  \item A more realistic degradation model than bicubic degradation model for SISR is proposed. It considers arbitrary blur kernels and enables to use existing deblurring methods for blur kernel estimation.
  \vspace{-0.15cm}
  \item A deep plug-and-play super-resolution framework is proposed to solve SISR with the new degradation model. DPSR is applicable beyond bicubic degradation and can handle LR images with arbitrary blur kernels.
  \vspace{-0.15cm}
  \item The proposed DPSR is well-principled as the iterative scheme aims to solve the new degradation induced energy function.
  \vspace{-0.15cm}
  \item The proposed DPSR extends existing plug-and-play framework, showing that the plug-and-play prior for SISR is not limited to Gaussian denoiser.
\end{itemize}

\section{Related work}

\subsection{DNN-based SISR}
\paragraph{\emph{1) Bicubic degradation}.}
The first DNN-based SISR method, termed as SRCNN~\cite{dong2014learning}, employs a relatively shallow network and
follows the previous SISR methods such as A+~\cite{timofte2014a+} and ScSR~\cite{yang2008image} to synthesize LR images with bicubic interpolation.
Since then, by fixing the degradation model to bicubic degradation,
some researchers began to improve SISR performance via DNN from different aspects, including
PSNR and SSIM values, efficiency, and perceptual visual quality at a large scale factor.

To improve the SISR performance in terms of PSNR and SSIM, the very deep super-resolution (VDSR) network proposed by Kim~\etal~\cite{kim2015accurate}
shows that the most direct way is to increase the network depth.
However, VDSR operates on the bicubicly interpolated LR images which hinders the efficiency.
To that end, the FSRCNN~\cite{dong2016accelerating} and ESPCN~\cite{shi2016real} are proposed to directly manipulate the LR
input and adopt an upscaling operation at the end of the network.
Considering the fact that the visual results tend to be oversmoothed at a large scale factor (\eg, 4),
VGG~\cite{simonyan2014very} loss and generative adversarial network (GAN)~\cite{goodfellow2014generative} loss are utilized to
improve the perceptual visual quality in~\cite{ledig2016photo,sajjadi2017enhancenet,wang2018esrgan}.

While achieving great success on bicubic degradation~\cite{zhang2018residual,lim2017enhanced,plotz2018neural}, these methods perform poorly on most of real images due to the mismatch of degradation models.

\vspace{-0.4cm}
\paragraph{\emph{2) Beyond bicubic degradation}.}
In~\cite{efrat2013accurate}, the authors pointed out that an accurate estimate of the blur kernel is more important than sophisticated image prior.
Since then, several attempts have been made to tackle with LR images beyond bicubic degradation.
Zhang~\etal~\cite{zhang2017learning} proposed a plug-and-play framework (IRCNN) to solve the energy function induced by Eqn.~\eqref{degradation:eq1}.
Although in theory IRCNN can handle arbitrary blur kernel (please refer to~\cite{chan2016plug}), the blur kernel of such degradation model in practice is difficult to estimate.
Zhang~\etal~\cite{zhang2018learning} proposed a general DNN-based solution (SRMD) which takes two key degradation parameters as input.
However, SRMD only considers the Gaussian blur kernels.
In~\cite{shocher2018zero}, the authors proposed a zero-shot super-resolution (ZSSR) method which trains image-specific DNN on the testing LR image and can also
take the degradation parameters, such as the estimated blur kernel, to improve the performance.
While showing impressive results for LR image with internal repetitive structures, ZSSR is less effective for severely blurred LR image.

As discussed, the above methods have two main drawbacks. First, they have difficulty in blur kernel estimation.
Second, they are generally designed for Gaussian-like blur kernel and thus cannot effectively handle severely blurred LR image.
It should be noted that a deep blind SISR method for motion blur is proposed in~\cite{zhang2018gated}. However, it has limited ability to handle the distortion of arbitrary blur kernels.

\subsection{Plug-and-play image restoration}
The plug-and-play image restoration which was first introduced in~\cite{danielyan2010image,zoran2011learning,venkatakrishnan2013plug}
has attracted significant attentions due to its flexibility and effectiveness in handling various inverse problems.
Its main idea is to unroll the energy function by variable splitting technique and
replace the prior associated subproblem by any off-the-shelf Gaussian denoiser.
Different from traditional image restoration methods which employ hand-crafted
image priors, it can implicitly define the plug-and-play prior by the denoiser.
Remarkably, the denoiser can be learned by DNN with large capability which would give rise to promising performance.

During the past few years, a flurry of plug-and-play works have been developed from the following aspects:
1) different variable splitting algorithms, such as half-quadratic splitting (HQS) algorithm~\cite{afonso2010fast}, alternating direction method of multipliers (ADMM) algorithm~\cite{boyd2011distributed}, FISTA~\cite{beck2009fast}, and primal-dual algorithm~\cite{chambolle2011first,ono2017primal};
2) different applications, such as Poisson denoising~\cite{rond2016poisson}, demosaicking~\cite{heide2014flexisp}, deblurring~\cite{tirer2018image}, super-resolution~\cite{chan2016plug,brifman2016turning,zhang2017learning,jiang2018deep}, and inpainting~\cite{meinhardt2017learning};
3) different types of denoiser priors, such as BM3D~\cite{dabov2007image,egiazarian2015single}, DNN-based denoisers~\cite{zhang2017beyond,bigdeli2017deep} and their combinations~\cite{gu2018integrating};
and 4) theoretical analysis on the convergence from the aspect of fixed point~\cite{chan2016plug,liu2018proximal,liu2019learning} and Nash equilibrium~\cite{buzzard2018plug,reehorst2018regularization,danielyan2012bm3d}.

To the best of our knowledge, existing plug-and-play image restoration methods mostly treat the Gaussian denoiser as the prior.
We will show that, for the application of plug-and-play SISR, the prior is not limited to Gaussian denoiser.
Instead, a simple super-resolver prior can be employed to solve a much more complex SISR problem.

\section{Method}

\subsection{New degradation model}
In order to ease the blur kernel estimation, we propose the following degradation model
\begin{equation}\label{eq_degradation3}
  \mathbf{y} = (\mathbf{x}\!\downarrow_s)\otimes\mathbf{k} + ~\mathbf{n},
\end{equation}
where $\downarrow_s$ is the bicubic downsampler with scale factor $s$.
Simply speaking, Eqn.~\eqref{eq_degradation3} conveys that the LR image $\mathbf{y}$
is a bicubicly downsampled, blurred and noisy version of a clean HR image $\mathbf{x}$.

Since existing methods widely use bicubic downsampler to synthesize or augment LR image,
it is a reasonable assumption that bicubicly downsampled HR image (\ie, $\mathbf{x}\!\downarrow_s$) is also a clean image.
Following this assumption, Eqn.~\eqref{eq_degradation3} actually corresponds to a deblurring problem followed by a SISR problem with bicubic degradation.
Thus, we can fully employ existing well-studied deblurring methods to estimate $\mathbf{k}$.
Clearly, this is a distinctive advantage over the degradation model given by Eqn.~\eqref{degradation:eq1}.

Once the degradation model is defined, the next step is to formulate the energy function.
According to Maximum A Posteriori (MAP) probability, the energy function is formally given by
\begin{equation}\label{eq_map}
  \min\displaystyle_{\mathbf{x}} \frac{1}{2\sigma^2}\|\mathbf{y} - (\mathbf{x}\!\downarrow_s) \otimes \mathbf{k}\|^2 + \lambda \Phi(\mathbf{x}),
\end{equation}
where $\frac{1}{2\sigma^2}\|\mathbf{y} - (\mathbf{x}\!\downarrow_s) \otimes \mathbf{k}\|^2$ is the data fidelity (likelihood) term\footnote{In order to facilitate and clarify the parameter setting, we emphasize that, from the Bayesian viewpoint, the data fidelity term should be $\frac{1}{2\sigma^2}\|\mathbf{y} - (\mathbf{x}\!\downarrow_s) \otimes \mathbf{k}\|^2$ rather than $\frac{1}{2}\|\mathbf{y} - (\mathbf{x}\!\downarrow_s) \otimes \mathbf{k}\|^2$.} determined by the degradation model of Eqn.~\eqref{eq_degradation3},
$\Phi(\mathbf{x})$ is the regularization (prior) term, and $\lambda$ is the regularization parameter.
For discriminative learning methods,
their inference models actually correspond to an energy function where the degradation model is implicitly defined by the training LR and HR pairs.
This explains why existing DNN-based SISR methods trained on bicubic degradation perform poorly for real images.

\subsection{Deep plug-and-play SISR}\label{section:deepplug}
To solve Eqn.~\eqref{eq_map}, we first adopt the variable splitting technique to introduce an auxiliary variable $\mathbf{z}$, leading to the following equivalent constrained optimization formulation:
\begin{multline}\label{eq_variablesplitting}
\quad \quad \hat{\mathbf{x}} = \arg\min\displaystyle_{\mathbf{x}}\frac{1}{2\sigma^2}\|\mathbf{y} - \mathbf{z}\otimes\mathbf{k}\|^2 + \lambda\Phi(\mathbf{x}), \\
  \text{subject to} \quad \mathbf{z} = \mathbf{x}\!\downarrow_s.  \quad
\end{multline}
We then address Eqn.~\eqref{eq_variablesplitting} with half quadratic splitting (HQS) algorithm.
Note that other algorithms such as ADMM can also be exploited. We use HQS for its simplicity.

Typically, HQS tackles with Eqn.~\eqref{eq_variablesplitting} by minimizing the following problem which involves an additional quadratic penalty term
\begin{equation}\label{eq_mapq}
  \mathcal{L}_\mu(\mathbf{x}, \mathbf{z}) = \frac{1}{2\sigma^2}\|\mathbf{y} - \mathbf{z}\otimes\mathbf{k}\|^2 + \lambda\Phi(\mathbf{x}) + \frac{\mu}{2}\|\mathbf{z} - \mathbf{x}\!\downarrow_s\!\|^2,
\end{equation}
where $\mu$ is the penalty parameter, and a very large $\mu$ will enforce $\mathbf{z}$ approximately equals to $\mathbf{x}\!\downarrow_s$.
Usually, $\mu$ varies in a non-descending order during the following iterative solution to Eqn.~\eqref{eq_mapq}
\begin{numcases}{}
\!\!\!\mathbf{z}_{k+1}=\arg\min\displaystyle_{\mathbf{z}}\|\mathbf{y}-\mathbf{z}\!\otimes\!\mathbf{k}\|^2 + \mu\sigma^2\|\mathbf{z}-\mathbf{x}_{k}\!\downarrow_s\!\|^2, \quad \label{eq_s1}\\
\!\!\!\mathbf{x}_{k+1}=\arg\min\displaystyle_{\mathbf{x}}\frac{\mu}{2}\|\mathbf{z}_{k+1} - \mathbf{x}\!\downarrow_s\!\|^2 + \lambda\Phi(\mathbf{x}). \label{eq_s2}
\end{numcases}
It can be seen that Eqn.~\eqref{eq_s1} and Eqn.~\eqref{eq_s2} are alternating minimization problems with respect to $\mathbf{z}$ and $\mathbf{x}$, respectively.
In particular, by assuming the convolution is carried out with circular boundary conditions, Eqn.~\eqref{eq_s1} has a fast closed-form solution
\begin{equation}\label{eq_fft}
  \mathbf{z}_{k+1} = \mathcal{F}^{-1}\left(\frac{\overline{\mathcal{F}(\mathbf{k})}\mathcal{F}(\mathbf{y})+\mu\sigma^2\mathcal{F}(\mathbf{x}_{k}\!\downarrow_s) }{\overline{\mathcal{F}(\mathbf{k})} \mathcal{F}(\mathbf{k})+\mu\sigma^2}\right),
\end{equation}
where $\mathcal{F}(\cdot)$ and $\mathcal{F}^{-1}(\cdot)$ denote the Fast Fourier Transform
(FFT) and inverse FFT, $\overline{\mathcal{F}(\cdot)}$ denotes complex
conjugate of $\mathcal{F}(\cdot)$.

To analyze Eqn.~\eqref{eq_s2} from a Bayesian perspective, we rewrite it as follows
\begin{equation}\label{eq_srproblem}
  \mathbf{x}_{k+1}=\arg\min\displaystyle_{\mathbf{x}}\frac{1}{2(\sqrt{1/\mu})^2}\|\mathbf{z}_{k+1} - \mathbf{x}\!\downarrow_s\!\|^2 + \lambda\Phi(\mathbf{x}).
\end{equation}
Clearly, Eqn.~\eqref{eq_srproblem} corresponds to super-resolving $\mathbf{z}_{k+1}$ with a scale factor $s$ by assuming $\mathbf{z}_{k+1}$ is bicubicly downsampled from an HR image $\mathbf{x}$, and then corrupted by AWGN with noise level $\sqrt{1/\mu}$.
From another viewpoint, Eqn.~\eqref{eq_srproblem} solves a super-resolution problem with the following simple bicubic degradation model
\begin{equation}\label{eq_degradation4}
\mathbf{y} = \mathbf{x}\!\downarrow_s + ~\mathbf{n}.
\end{equation}
As a result, one can plug DNN-based super-resolver trained on the widely-used bicubic degradation with certain noise levels to replace Eqn.~\eqref{eq_srproblem}.
For brevity, Eqn.~\eqref{eq_s2} and Eqn.~\eqref{eq_srproblem} can be further rewritten as
\begin{equation}\label{eq_srfinal}
  \mathbf{x}_{k+1} = \mathcal{SR}(\mathbf{z}_{k+1}, s, \sqrt{1/\mu}).
\end{equation}

Since the prior term $\Phi(\mathbf{x})$ is implicitly defined in $\mathcal{SR}(\cdot)$, we refer to it as super-resolver prior.

So far, we have seen that the two sub-problems given by Eqn.~\eqref{eq_s1} and Eqn.~\eqref{eq_s2} are relatively easy to solve.
In fact, they also have clear interpretation.
On the one hand, since the blur kernel $\mathbf{k}$ is only involved in the closed-form solution, Eqn.~\eqref{eq_s1} addresses the distortion of blur.
In other words, it pulls the current estimation to a less blurry one.
On the other hand, Eqn.~\eqref{eq_s2} maps the less blurry image to a more clean HR image.
After several alternating iterations,
it is expected that the final reconstructed HR image contains no blur and noise.

\subsection{Deep super-resolver prior}
In order to take advantage of the merits of DNN, we need to specify the super-resolver network which should take the noise level as input according to Eqn.~\eqref{eq_srfinal}.
Inspired by~\cite{gharbi2016deep,zhang2018ffdnet}, we only need to modify most of the existing DNN-based super-resolvers by taking an additional noise level map as input.
Alternatively, one can directly adopt SRMD as the super-resolver prior because its input already contains the noise level map.

Since SRResNet~\cite{ledig2016photo} is a well-known DNN-based super-resolver, in this paper we propose a modified SRResNet, namely SRResNet+, to plug in the proposed DPSR framework.
SRResNet+ differs from SRResNet in several aspects.
First, SRResNet+ additionally takes a noise level map $\mathbf{M}$ as input.
Second, SRResNet+ increases the number of feature maps from 64 to 96.
Third, SRResNet+ removes the batch normalization layer~\cite{ioffe2015batch} as suggested in~\cite{wang2018esrgan}.

Before training a separate SRResNet+ model for each scale factor, we need to synthesize the LR image and its noise level map from a given HR image.
According to the degradation model given by Eqn.~\eqref{eq_degradation4},
the LR image is bicubicly downsampled from an HR image, and then corrupted by AWGN with a noise level $\sigma$ from predefined noise level range.
For the corresponding noise level map, it has the same spatial size of LR image and all the elements are $\sigma$. Following~\cite{zhang2018learning}, we set the noise level range to $[0, 50]$.
For the HR images, we choose the 800 training images from DIV2K dataset~\cite{agustsson2017ntire}.

We adopt Adam algorithm~\cite{kingma2014adam} to optimize SRResNet+ by minimizing the $\ell_1$ loss function. 
The leaning rate starts from $10^{-4}$, then decreases by half every $5\times10^5$ iterations and finally ends once it is smaller than $10^{-7}$.
The mini-batch size is set to 16. The patch size of LR input is set to 48$\times$48.
The rotation and flip based data augmentation is performed during training.
We train the models with PyTorch on a single GTX 1080 Ti GPU.

Since this work mainly focuses on SISR with arbitrary blur kernels. We omit the comparison between SRResNet+ and other methods on bicubic degradation.
As a simple comparison, SRResNet+ can outperform SRResNet~\cite{ledig2016photo} by an average PSNR gain of 0.15dB on Set5~\cite{bevilacqua2012low}.

\subsection{Comparison with related methods}
In this section, we emphasize the fundamental differences between the proposed DPSR and several closely related
DNN-based methods.

\vspace{-0.4cm}
\paragraph{\emph{1) Cascaded deblurring and SISR.}}
To super-resolve LR image with arbitrary blur kernels, a heuristic method is to perform deblurring first and then super-resolve the deblurred LR image.
However, such a cascaded two-step method suffers from the drawback that the perturbation error of
the first step would be amplified at the second step.
On the contrary, DPSR optimizes the energy function given by Eqn.~\eqref{eq_map} in an iterative manner.
Thus, DPSR tends to deliver better performance.

\vspace{-0.4cm}
\paragraph{\emph{2) Fine-tuned SISR model with more training data.}}
Perhaps the most straightforward way is to fine-tune existing bicubic degradation based SISR models with more training data generated by the new degradation model (\ie, Eqn.~\eqref{eq_degradation3}),
resulting in the so-called blind SISR.
However, the performance of such methods deteriorates seriously especially when large complex blur kernels are considered, possibly because the distortion of blur would further aggravate the pixel-wise average problem~\cite{ledig2016photo}. As for DPSR, it takes the blur kernel as input and can effectively handle the distortion of blur via Eqn.~\eqref{eq_fft}.

\vspace{-0.4cm}
\paragraph{\emph{3) Extended SRMD or DPSR with end-to-end training.}}
Inspired by SRMD~\cite{zhang2018learning}, one may attempt to extend it by considering arbitrary blur kernels.
However, it is difficult to sample enough blur kernels to cover the large kernel space. In addition, it would require a large amount of time to train a reliable model.
By contrast, DPSR only needs to train the models on the bicubic degradation, thus it involves much less training time.
Furthermore, while SRMD can effectively handle the simple Gaussian kernels of size $15\times15$ with many successive convolutional layers, it loses effectiveness to deal with large complex blur kernels. Instead, DPSR adopts a more concise and specialized modular by FFT via Eqn.~\eqref{eq_fft} to eliminate the distortion of blur.
Alternatively, one may take advantage of the structure benefits of DPSR and resort to jointly training DPSR in an end-to-end manner.
However, we leave this to our future work.

From the above discussions, we can conclude that our DPSR is well-principled, structurally simple, highly interpretable and involves less training.

\section{Experiments}

\subsection{Synthetic LR images}
Following the common setting in most of image restoration literature, we use synthetic data with ground-truth to quantitatively analyze the proposed DPSR, as well as making a relatively fair comparison with other competing methods.

\begin{figure}[!htbp]\footnotesize
\scriptsize{
\begin{center}
\subfigure[Gaussian]
{\includegraphics[width=0.15\textwidth]{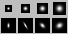}}
\hspace{0.1cm}
\subfigure[Motion]
{\includegraphics[width=0.15\textwidth]{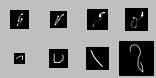}}
\hspace{0.1cm}
\subfigure[Disk]
{\includegraphics[width=0.15\textwidth]{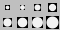}}
\caption{Examples of (a) Gaussian blur kernels, (b) motion blur kernels and (c) disk blur kernels.}\label{fig_kernel}
\end{center}}\vspace{-0.4cm}
\end{figure}

\vspace{-0.4cm}
\paragraph{Blur kernel.}
For the sake of thoroughly evaluating the effectiveness of the proposed DPSR for arbitrary blur kernels, we consider three types of widely-used blur kernels, including Gaussian blur kernels,
motion blur kernels, and disk (out-of-focus) blur kernels~\cite{chan2011single,xu2014deep}. The specifications of the blur kernels are given in Table~\ref{table_kernels}.
Some kernel examples are shown in Fig.~\ref{fig_kernel}.
Note that the kernel sizes range from 5$\times$5 to 35$\times$35.
As shown in Table~\ref{table_psnr}, we further consider Gaussian noise with two different noise levels, \ie, 2.55 (1\%) and 7.65 (3\%), for scale factor 3.

\begin{table}[!htbp]\footnotesize\vspace{-0.2cm}
\caption{Three different types of blur kernels.}
\center
\begin{tabular}{p{1.1cm}<{\centering}|p{0.4cm}<{\centering}|p{5.5cm}}
  \shline
   Type & $\#$ & \quad \quad \quad \quad \quad \quad \quad Specification \\ \hline
  \multirow{4}{*}{Gaussian} & \multirow{4}{*}{16}  & 8 isotropic Gaussian kernels with standard deviations uniformly sampled from the interval [0.6, 2], and 8 selected anisotropic Gaussian blur kernels from~\cite{zhang2018learning}.    \\\hline
  \multirow{4}{*}{Motion} & \multirow{4}{*}{32}    & 8 blur kernels from~\cite{freeman2009understanding} and their augmented 8 kernels by random rotation and flip; and 16 realistic-looking motion blur kernels generated by the released code of~\cite{boracchiFoiTIP12a}. \\\hline
  \multirow{4}{*}{Disk} & \multirow{4}{*}{8} & 8 disk kernels with radius uniformly sampled from the interval [1.8, 6]. They are generated by matlab function \texttt{fspecial('disk',r)}, where \texttt{r} is the radius. \\
  \shline
\end{tabular}
\label{table_kernels}
\end{table}\vspace{-0.9cm}

\begin{table*}[!htbp]\footnotesize 
\caption{Average PSNR and SSIM results of different methods for different degradation settings on the color BSD68 dataset~\cite{MartinFTM01,roth2009fields,zhang2017beyond}. The best two results are highlighted in \textcolor[rgb]{1.00,0.00,0.00}{red} and \textcolor[rgb]{0.00,0.00,1.00}{blue} colors, respectively.} \vspace{-0.25cm}
\center
\begin{tabular}{p{0.7cm}<{\centering}|p{0.9cm}<{\centering}|p{0.6cm}<{\centering}|p{1.3cm}<{\centering}|p{1.3cm}<{\centering}|p{1.3cm}<{\centering}|p{1.3cm}<{\centering}|p{1.3cm}<{\centering}|p{1.3cm}<{\centering}|p{1.3cm}<{\centering}|p{1.3cm}<{\centering}}
  \shline
  \multicolumn{3}{c|}{Degradation Setting} & \multicolumn{8}{c}{Methods} \\ \hline
 Scale & Kernel & Noise & \multirow{2}{*}{Bicubic} & \multirow{2}{*}{VDSR} & \multirow{2}{*}{RCAN} & IRCNN  & \footnotesize{DeblurGAN}& \multirow{2}{*}{GFN} & \multirow{2}{*}{ZSSR} & \multirow{2}{*}{DPSR}\\
 Factor & Type & Level &   & &   & +RCAN  & +RCAN &  &    &    \\ \hline\hline
  & Gaussian &  &  23.47/0.596 &23.36/0.589  &23.59/0.603  & \textcolor{blue}{24.79}/\textcolor{blue}{0.629}  & 19.36/0.400 & --  & 21.44/0.542  & \textcolor{red}{27.28}/\textcolor{red}{0.763}\\
 $\times$2 & Motion & 0 &  19.84/0.449 &19.86/.451  & 19.82/0.448 & \textcolor{blue}{28.54}/\textcolor{blue}{0.806}& 17.46/0.268  & --  & 17.99/0.367 &\textcolor{red}{30.05}/\textcolor{red}{0.869}\\
  & Disk &  &  21.85/0.507  &21.86/0.508 & 21.85/0.507   & \textcolor{blue}{25.48}/\textcolor{blue}{0.671} & 19.33/0.370 & --  & 21.25/0.490 &\textcolor{red}{28.61}/\textcolor{red}{0.816}\\\hline

    & Gaussian&  & 22.11/0.526 &22.03/0.520 & 22.20/0.532 &  \textcolor{blue}{23.52}/\textcolor{blue}{0.566}  & 18.18/0.347  & --  &  18.97/0.442  & \textcolor{red}{25.22}/\textcolor{red}{0.665}\\
 & Motion  & 0 & 18.83/0.424  &18.84/0.424 & 18.81/0.422  &  \textcolor{blue}{25.88}/\textcolor{blue}{0.699} & 16.25/0.228  & -- & 16.80/0.348 & \textcolor{red}{27.22}/\textcolor{red}{0.769}\\
  & Disk &  &  20.70/0.464 &20.70/0.465  & 20.69/0.464  & \textcolor{blue}{23.82}/\textcolor{blue}{0.594}  & 18.28/0.336  & --  & 19.05/0.430  &\textcolor{red}{26.19}/\textcolor{red}{0.716}\\\cline{2-11}

    &Gaussian & \multirow{2}{*}{2.55} & 22.05/0.513 &21.95/0.507  & 22.05/0.495  &  \textcolor{blue}{22.95}/\textcolor{blue}{0.540}  &  17.64/0.195  & --   & 20.52/0.477   & \textcolor{red}{23.92}/\textcolor{red}{0.595}   \\
 $\times$3 & Motion  & \multirow{2}{*}{(1\%)} & 18.80/0.412  &18.81/0.413  & 18.74/0.390 & \textcolor{blue}{24.18}/\textcolor{blue}{0.604} & 15.55/0.122 & --  & 18.19/0.402 & \textcolor{red}{24.94}/\textcolor{red}{0.648} \\
  & Disk& &  20.65/0.452  &20.66/0.453  & 20.59/0.429  & \textcolor{blue}{22.33}/\textcolor{blue}{0.517}  & 16.74/0.142  & --  &  19.88/0.441 & \textcolor{red}{23.44}/\textcolor{red}{0.567} \\\cline{2-11}

      &Gaussian & \multirow{2}{*}{7.65} & 21.60/0.436 &21.54/0.433  &21.01/0.342 & \textcolor{blue}{22.07}/\textcolor{blue}{0.463}  &  14.93/0.094  & --   & 20.22/0.407 & \textcolor{red}{23.17}/\textcolor{red}{0.557} \\
 & Motion  & \multirow{2}{*}{(3\%)} & 18.58/0.346  & 18.60/0.350  & 18.21/0.259  & \textcolor{blue}{22.47}/\textcolor{blue}{0.514} & 13.91/0.065  & -- & 18.07/0.340 & \textcolor{red}{23.28}/\textcolor{red}{0.564}  \\
  &Disk &  &  20.32/0.380  &20.34/0.384  & 19.81/0.288 & \textcolor{blue}{21.19}/\textcolor{blue}{0.483}  & 14.43/0.071 & --  & 19.64/0.373 & \textcolor{red}{22.25}/\textcolor{red}{0.515}\\\hline

    &Gaussian &  &  21.18/0.486 &21.11/0.482  & 21.27/0.491  & \textcolor{blue}{22.61}/\textcolor{blue}{0.526}  &  17.52/0.329  & 21.66/0.507   & 16.83/0.380  &\textcolor{red}{24.04}/\textcolor{red}{0.604} \\
 $\times$4 &Motion &0 &  18.10/0.407  &18.11/0.408 & 18.08/0.406& \textcolor{blue}{24.42}/\textcolor{blue}{0.625}  & 15.68/0.225 & 17.79/0.398& 15.58/0.329 &\textcolor{red}{25.69}/\textcolor{red}{0.695}\\
  &Disk & &  19.86/0.439  &19.86/0.440  & 19.85/0.439   &  \textcolor{blue}{22.72}/\textcolor{blue}{0.540} & 17.59/0.321  & 20.38/0.460  & 16.91/0.392 &\textcolor{red}{24.84}/\textcolor{red}{0.649}\\\shline

\end{tabular}
\label{table_psnr}\vspace{-0.1cm}
\end{table*}

\vspace{-0.4cm}
\paragraph{Parameter setting.}
In the alternating iterations between Eqn.~\eqref{eq_s1} and Eqn.~\eqref{eq_s2}, we need to set $\lambda$ and tune $\mu$ to obtain a satisfying performance.
Setting such parameters has been considered as a non-trivial task~\cite{romano2017little}.
However, the parameter setting of DPSR is generally easy with the following two principles.
First, since $\lambda$ is fixed and can be absorbed into $\sigma$, we can instead multiply $\sigma$ by a scalar $\sqrt{\lambda}$ and therefore ignore the $\lambda$ in Eqn.~\eqref{eq_s2}.
Second, since $\mu$ has a non-descending order during iterations, we can instead set the $\sqrt{1/\mu}$ from Eqn.~\eqref{eq_srfinal} with a non-ascending order
to indirectly determine $\mu$ in each iteration.
Empirically, a good rule of thumb is to set $\lambda$ to $1/3$ and exponentially decrease $\sqrt{1/\mu}$  from $49$ to a small $\sigma$-dependent value (\eg, $\max(2.55, \sigma)$) for a total of 15 iterations.


\vspace{-0.4cm}
\paragraph{Compared methods.}
We compare the proposed DPSR with six methods, including two representative DNN-based methods for bicubic degradation (\ie, VDSR~\cite{kim2015accurate} and RCAN~\cite{zhang2018image}),
two cascaded deblurring and SISR methods (\ie, IRCNN+RCAN and DeblurGAN+RCAN), and two specially designed methods for blurry LR images (\ie, GFN~\cite{zhang2018gated} and ZSSR~\cite{shocher2018zero}).
To be specific, VDSR is the first very deep network for SISR; RCAN consists of more than 400 layers and achieves
state-of-the-art performance for bicubic degradation;
IRCNN is a plug-and-play method with deep denoiser prior that can handle non-blind image deblurring;
DeblurGAN~\cite{kupyn2018deblurgan} is a deep blind deblurring method based on generative adversarial network (GAN)~\cite{goodfellow2014generative};
GFN is a DNN-based method for joint blind motion deblurring and super-resolution;
ZSSR is an unsupervised DNN-based method that can super-resolve blurry and noisy LR images.
Note that IRCNN, ZSSR and DPSR can take the blur kernel and noise level as input.
For a fair comparison, we modify ZSSR to our new degradation model.

\vspace{-0.4cm}
\paragraph{Quantitative results.}
The PSNR and SSIM results of different methods for different degradation settings on the color BSD68 dataset~\cite{MartinFTM01,roth2009fields,zhang2017beyond} are shown in Table~\ref{table_psnr},
from which we have several observations. First, while RCAN outperforms VDSR by a large margin for bicubic degradation (see~\cite{zhang2018image}), it has comparable performance to VDSR and even bicubic interpolation for the complex degradations settings. Such phenomenon has also been reported in~\cite{shocher2018zero,zhang2018learning}.
Second, after a deblurring step by IRCNN, IRCNN+RCAN can significantly improve the PSNR and SSIM values.
Third, DeblurGAN+RCAN and GFN give rise to poor performance which may be attributed to the limited capability of successive convolutional layers in handling distortion of large complex blur.
Fourth, ZSSR is less effective for large complex blur kernels due to the lack of recurrence property of the blurry LR image.
Last, our DPSR achieves the best performance since it directly optimizes the energy function for the given degradation and can effectively handle the distortion of blur via Eqn.~\eqref{eq_fft}.

\begin{figure*}[!htbp]\footnotesize
    \hspace{-0.23cm}
    \begin{tabular}{c@{\extracolsep{0.23em}}c@{\extracolsep{0.23em}}c@{\extracolsep{0.23em}}c@{\extracolsep{0.23em}}c}
		\includegraphics[width=0.193\textwidth]{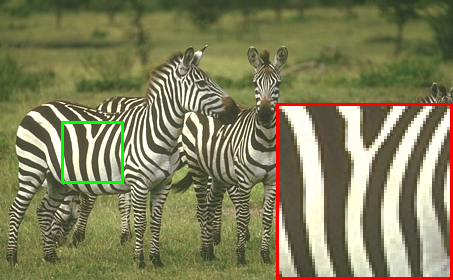}~
		&\includegraphics[width=0.193\textwidth]{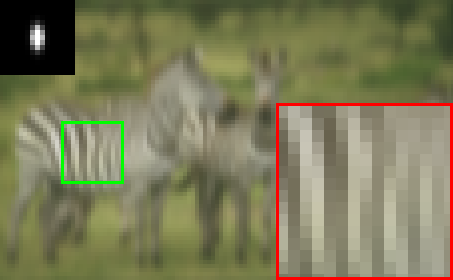}~
		&\includegraphics[width=0.193\textwidth]{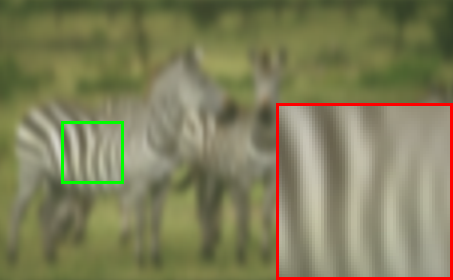}~
		&\includegraphics[width=0.193\textwidth]{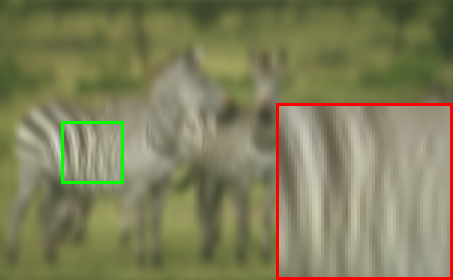}~
		&\includegraphics[width=0.193\textwidth]{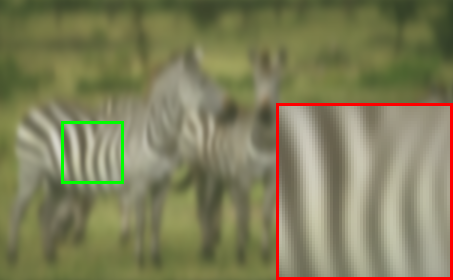}\\
        PSNR/SSIM&$\times$4&(17.25/0.406)&(17.12/0.397)&(17.37/0.416)\\
		(a) Ground-truth & (b) LR  & (c) Bicubic & (d) VDSR  & (e) RCAN \\
		\includegraphics[width=0.193\textwidth]{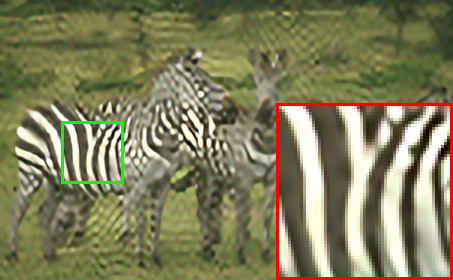}~
		&\includegraphics[width=0.193\textwidth]{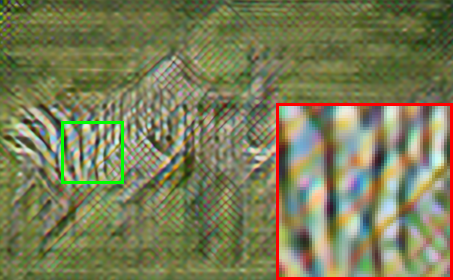}~
		&\includegraphics[width=0.193\textwidth]{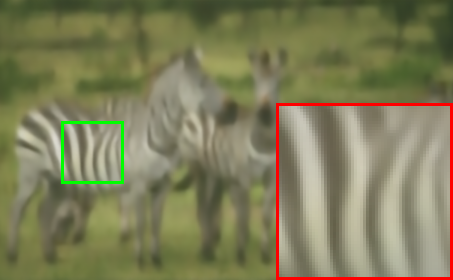}~
		&\includegraphics[width=0.193\textwidth]{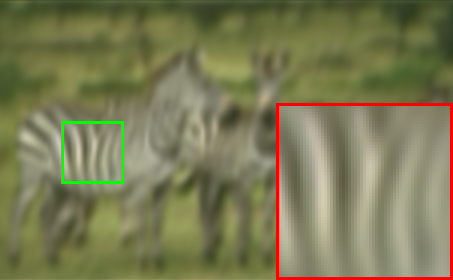}~
		&\includegraphics[width=0.193\textwidth]{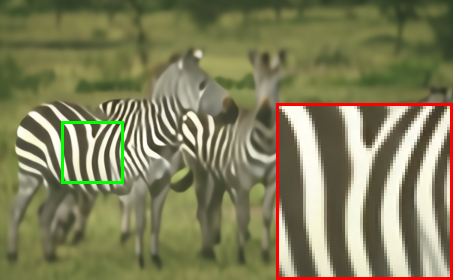}\\
		(18.28/0.472)&(15.29/0.227)&(17.58/0.437) &(16.61/0.374)& (19.44/0.577)\\
        (f) IRCNN+RCAN & (g) DeblurGAN+RCAN & (h) GFN & (i) ZSSR & (j) DPSR(ours)\\
	\end{tabular}
	\caption{The performance comparison of different methods for scale factor 4 on image ``\emph{253027}'' with Gaussian kernel. The blur kernel is shown on the upper-left corner of the LR image.}
	\label{fig_visual1}
\end{figure*}

\begin{figure*}[!htbp]\footnotesize 
\hspace{-0.23cm}
	\begin{tabular}{c@{\extracolsep{0.18em}}c@{\extracolsep{0.18em}}c@{\extracolsep{0.18em}}c@{\extracolsep{0.18em}}c@{\extracolsep{0.18em}}c}
        \includegraphics[width=0.16\textwidth]{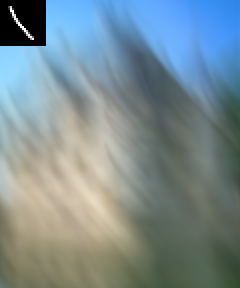}~
		&\includegraphics[width=0.16\textwidth]{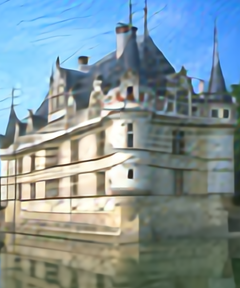}~
		&\includegraphics[width=0.16\textwidth]{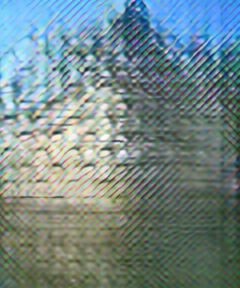}~
        &\includegraphics[width=0.16\textwidth]{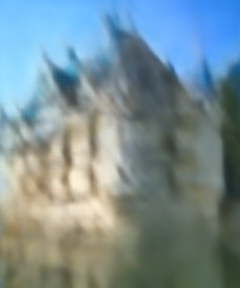}~
        &\includegraphics[width=0.16\textwidth]{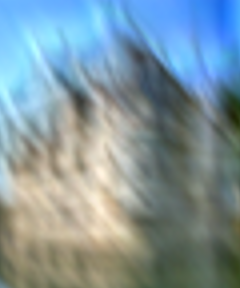}~
		&\includegraphics[width=0.16\textwidth]{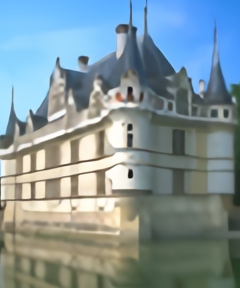}\\
		17.87/0.510 & 23.13/0.693  & 15.44/0.280  & 17.43/0.493  & 17.27/0.474 & 23.75/0.739\\
		\includegraphics[width=0.16\textwidth]{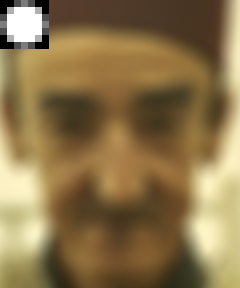}~
		&\includegraphics[width=0.16\textwidth]{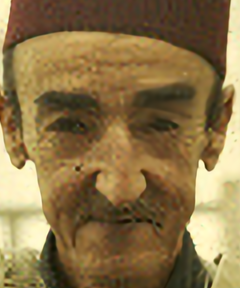}~
		&\includegraphics[width=0.16\textwidth]{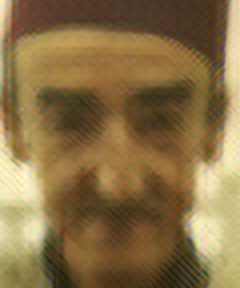}~
		&\includegraphics[width=0.16\textwidth]{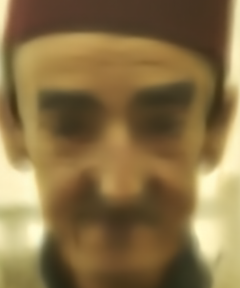}~
        &\includegraphics[width=0.16\textwidth]{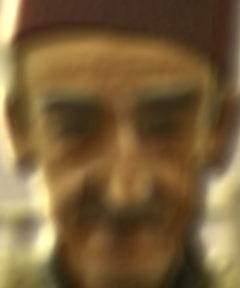}~
		&\includegraphics[width=0.16\textwidth]{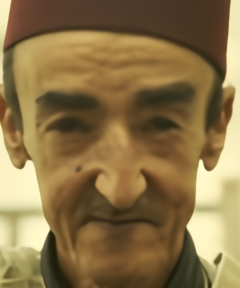}\\
		22.66/0.640 & 28.53/0.704  & 22.58/0.571  & 23.93/0.677  & 15.41/0.549 & 31.35/0.798\\
       (a) RCAN & (b) IRCNN+RCAN & (c) DeblurGAN+RCAN & (d) GFN & (e) ZSSR & (f) DPSR(ours)\\
	\end{tabular}
	\caption{The performance comparison of different methods for scale factor 4 on image ``\emph{102061}'' with motion kernel (first row) and image ``\emph{189080}'' with disk kernel (second row). The blur kernel is shown on the upper-left corner of the super-resolved image by RCAN.}
	\label{fig_visual2}
\end{figure*}

\begin{figure*}[!htbp]\footnotesize 
\hspace{-0.23cm}
	\begin{tabular}{c@{\extracolsep{0.18em}}c@{\extracolsep{0.18em}}c@{\extracolsep{0.18em}}c@{\extracolsep{0.18em}}c@{\extracolsep{0.18em}}c}
		\includegraphics[width=0.16\textwidth]{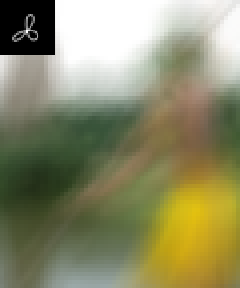}~
		&\includegraphics[width=0.16\textwidth]{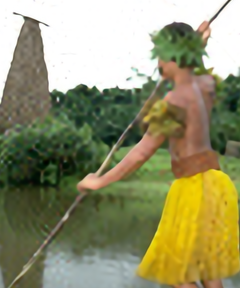}~
		&\includegraphics[width=0.16\textwidth]{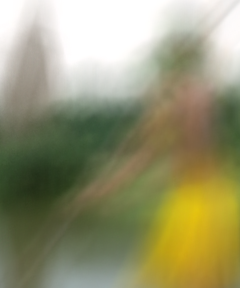}~
		&\includegraphics[width=0.16\textwidth]{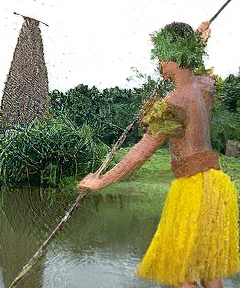}~
        &\includegraphics[width=0.16\textwidth]{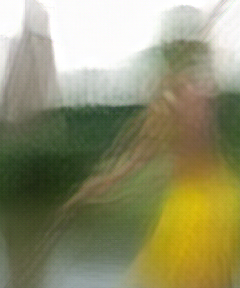}~
        &\includegraphics[width=0.16\textwidth]{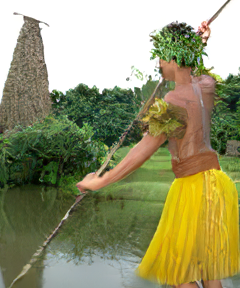}~\\
		\includegraphics[width=0.16\textwidth]{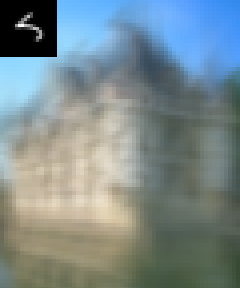}~
		&\includegraphics[width=0.16\textwidth]{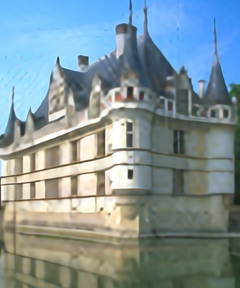}~
        &\includegraphics[width=0.16\textwidth]{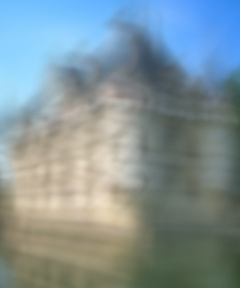}~
        &\includegraphics[width=0.16\textwidth]{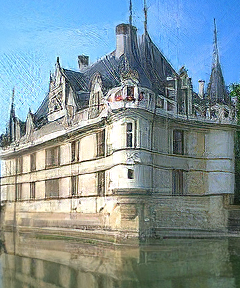}~
        &\includegraphics[width=0.16\textwidth]{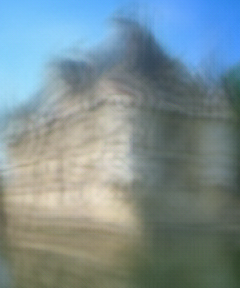}~
		&\includegraphics[width=0.16\textwidth]{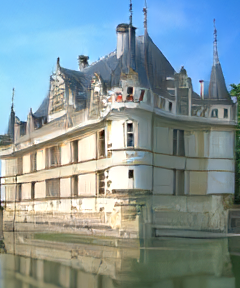}\\
		(a) LR ($\times4$)& (b) IRCNN+RCAN & (c) ESRGAN & (d) IRCNN+ESRGAN & (e) RCAN+DeblurGAN & (f) DPSRGAN(ours)\\
	\end{tabular}
	\caption{The performance comparison of different methods for scale factor 4 on image ``\emph{101087}'' (first row) and image ``\emph{102061}'' (second row). The blur kernel is shown on the upper-left corner of the LR image.}
	\label{fig:gan}\vspace{-0.5cm}
\end{figure*}

\paragraph{Visual results.}
Fig.~\ref{fig_visual1} shows the visual comparison of different methods for super-resolving LR image with Gaussian blur by a scale factor of 4.
As one can see, VDSR and RCAN undoubtedly produce unpleasant results due to the degradation mismatch.
DeblurGAN+RCAN generates very unpleasant artifacts and does not alleviate the blurriness compared to the LR image.
GFN does not perform well and ZSSR is less effective for super-resolving the very blurry LR image.
While IRCNN+RCAN produces better results than DeblurGAN+RCAN, GFN and ZSSR, it generates some noise-like artifacts which is possibly introduced by IRCNN and then amplified by RCAN.
In comparison, our DPSR produces the most visually pleasant results.

Fig.~\ref{fig_visual2} further shows the visual comparison on other two types of blur kernels.
It can be observed that DPSR consistently gives rise to the best visual results.
Particularly, although GFN can deal with the motion blur to some degree, its visual result is significantly inferior to those of IRCNN+RCAN and DPSR.
The underlying reason is that it has limited ability to blindly handle the distortion of blur via successive convolutional layers.
We note that other learning based methods (\eg,~\cite{tao2018scale}) also suffer from such problem.

\begin{figure}[!tbp]\footnotesize\vspace{0.2cm}
\hspace{-0.23cm}
	\begin{tabular}{c@{\extracolsep{0.23em}}c@{\extracolsep{0.23em}}}
		\includegraphics[width=0.23\textwidth]{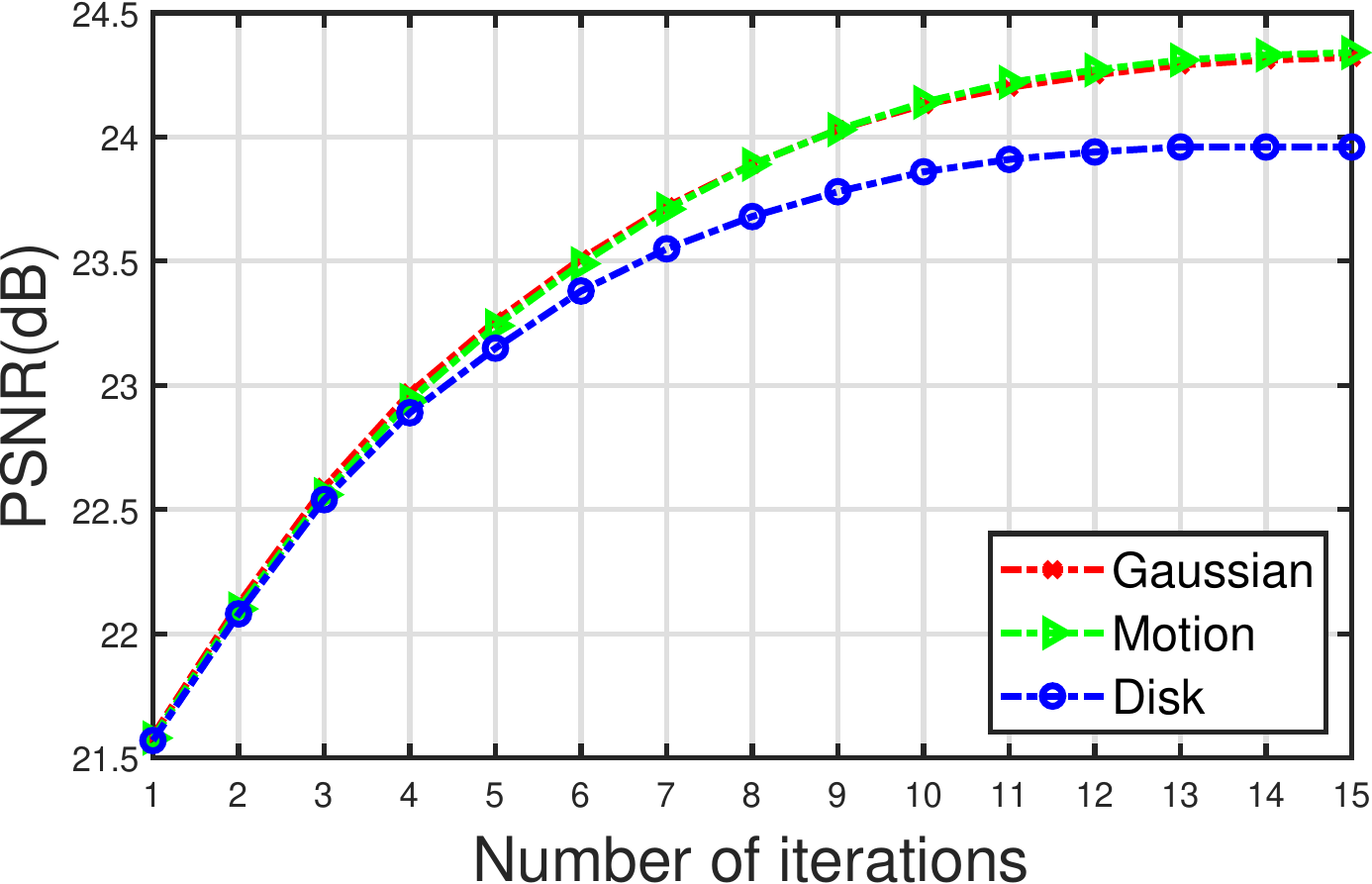}~
		&\includegraphics[width=0.23\textwidth]{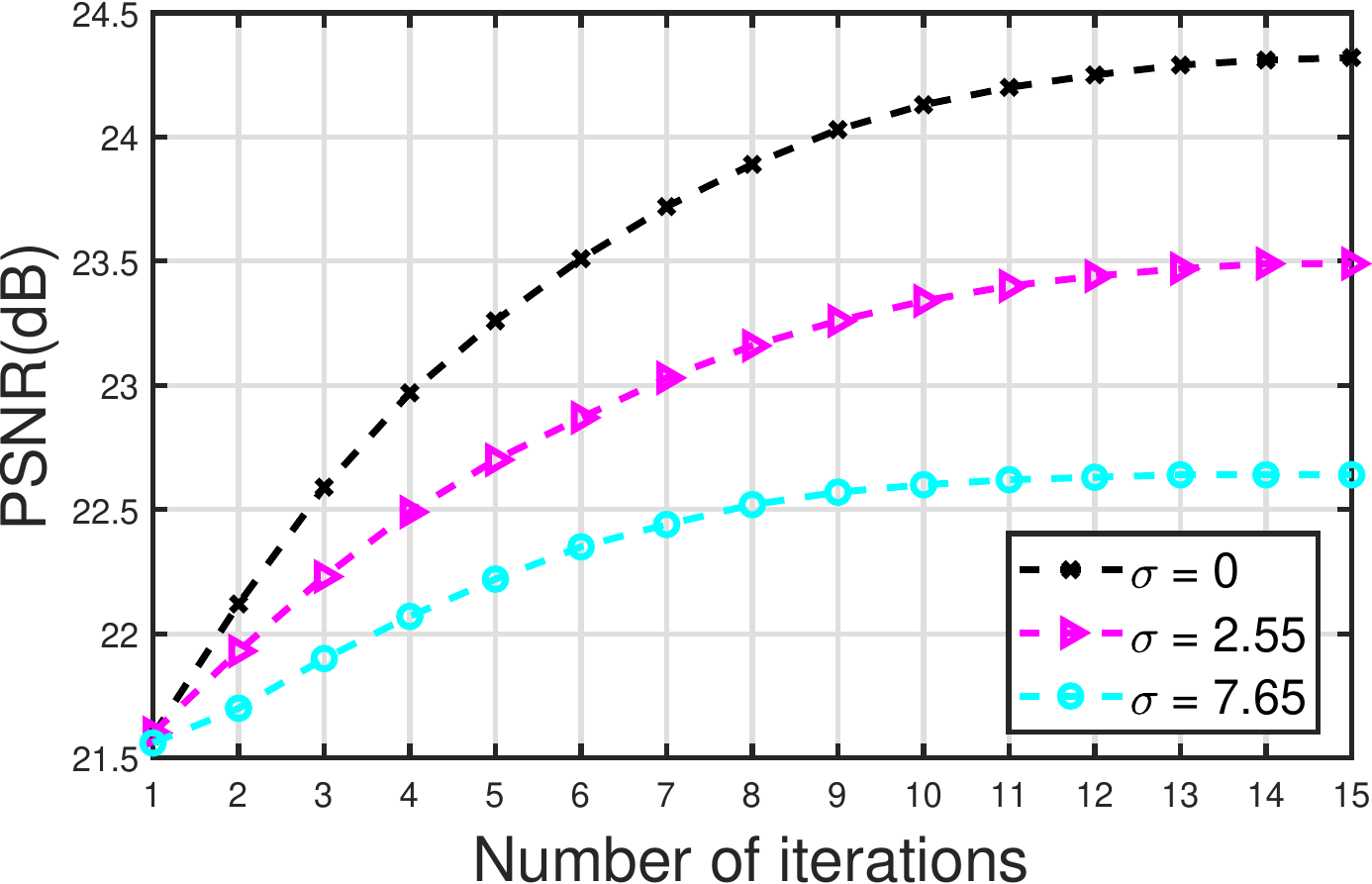}\\
		(a) Different types of kernels & (b) Different noise levels \\
	\end{tabular}
	\caption{Convergence curves of DPSR on image ``\emph{102061}'' with respect to different types of blur kernels and different noise levels.}
	\label{fig:convergence}\vspace{-0.4cm}
\end{figure}

\vspace{-0.4cm}
\paragraph{Convergence.}
Since our DPSR is a special case of plug-and-play framework for SISR, one may refer to~\cite{buzzard2018plug,reehorst2018regularization} to analyze the theoretical convergence.
In this paper, we only provide an empirical evidence to show the convergence of DPSR.
Fig.~\ref{fig:convergence} shows the convergence curves of the proposed DPSR on image ``\emph{102061}'' with respect to
different types of blur kernels and different noise levels.
In Fig.~\ref{fig:convergence}(a), we fix the noise level to 0 and choose the third kernel for each kernel type.
In Fig.~\ref{fig:convergence}(b), we fix the blur kernel to the third Gaussian kernel and choose three different noise levels, including 0, 2.55 and 7.65.
It can be observed that DPSR converges very fast.

It is worth pointing out, in order to reduce the computational cost, our DPSR does not necessarily need iterative solution for bicubic degradation
since our adopted super-resolver prior is already trained end-to-end on such degradation.
Obviously, this is also an advantage over existing plug-and-play SISR.
For the case of other degradations, the run time of our DPSR mainly depends on the total number of iterations.
On a single GPU, it takes about 1.8 seconds for DPSR to super-resolve an LR image of size 256$\times$256 by different scale factors.
As a comparison, ZSSR spends 12, 14 and 18 seconds for scale factor 2, 3 and 4, respectively.
In practice, one can tune the total number of iterations to balance the performance and speed.

\begin{figure*}[!tbp]\footnotesize
    \hspace{-0.23cm}
    \begin{tabular}{c@{\extracolsep{0.23em}}c@{\extracolsep{0.23em}}c@{\extracolsep{0.23em}}c@{\extracolsep{0.23em}}c}
		\includegraphics[width=0.193\textwidth]{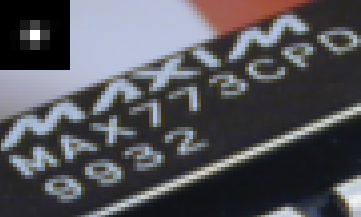}~
		&\includegraphics[width=0.193\textwidth]{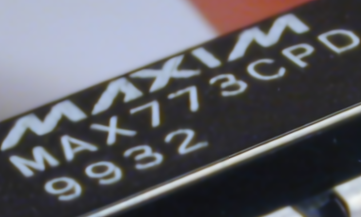}~
		&\includegraphics[width=0.193\textwidth]{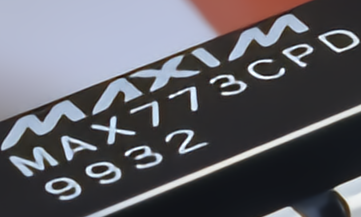}~
		&\includegraphics[width=0.193\textwidth]{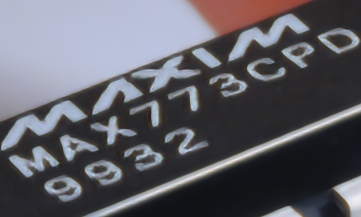}~
		&\includegraphics[width=0.193\textwidth]{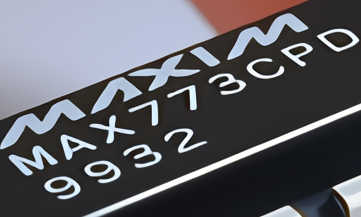}\\

		\includegraphics[width=0.193\textwidth]{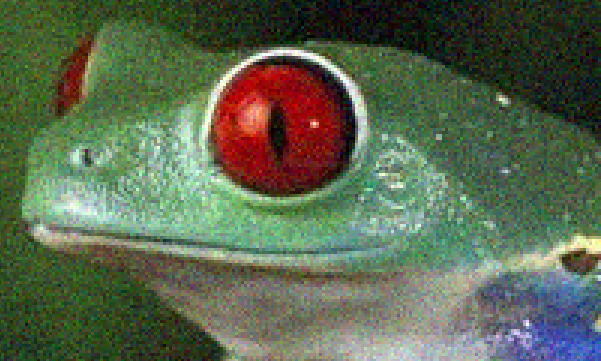}~
		&\includegraphics[width=0.193\textwidth]{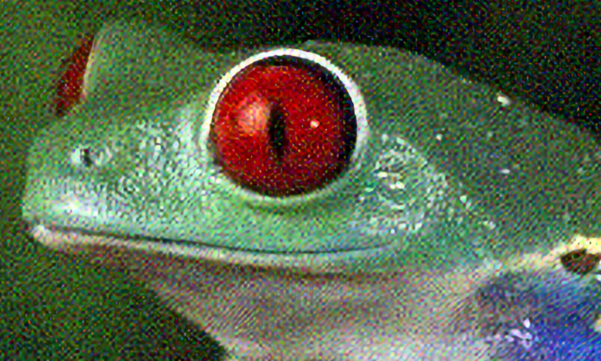}~
		&\includegraphics[width=0.193\textwidth]{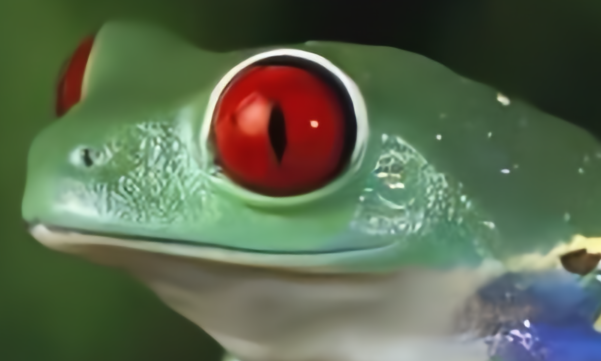}~
		&\includegraphics[width=0.193\textwidth]{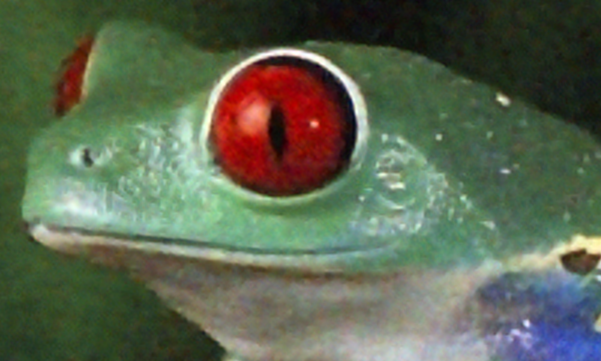}~
		&\includegraphics[width=0.193\textwidth]{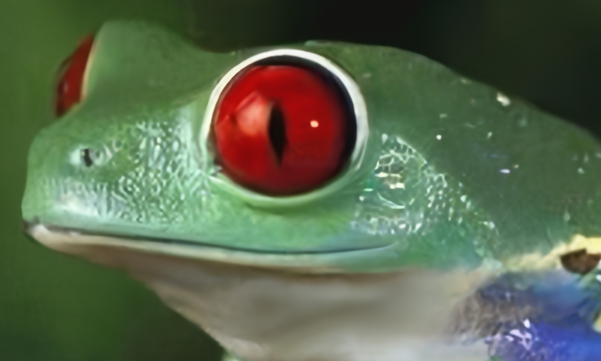}\\

        \includegraphics[width=0.193\textwidth]{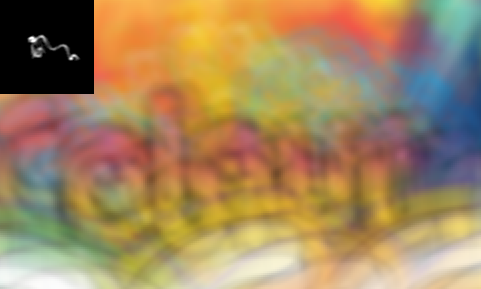}~
		&\includegraphics[width=0.193\textwidth]{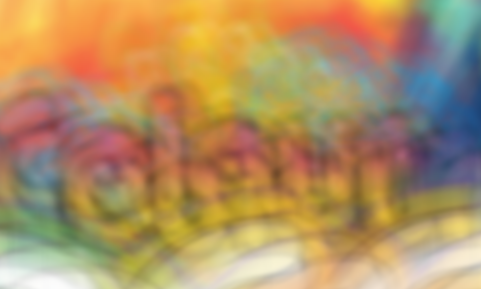}~
		&\includegraphics[width=0.193\textwidth]{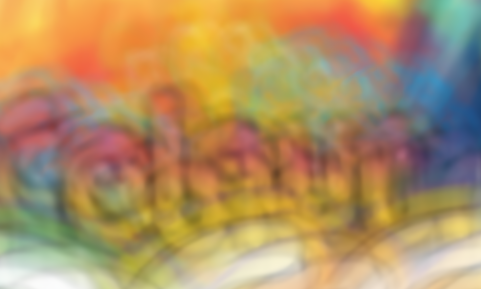}~
		&\includegraphics[width=0.193\textwidth]{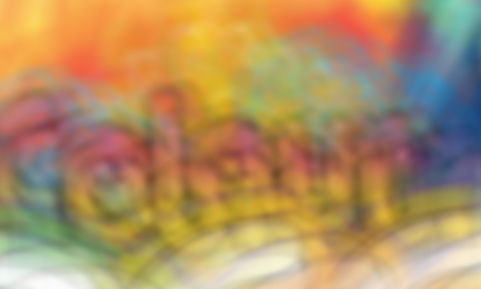}~
		&\includegraphics[width=0.193\textwidth]{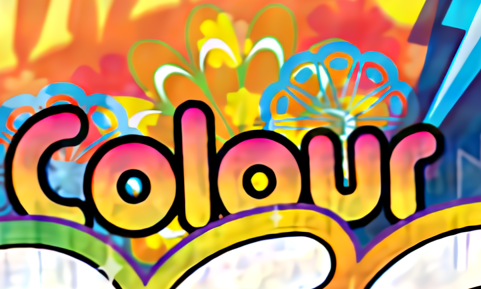}\\
        (a) LR & (b) RCAN  & (c) SRMD & (d) ZSSR  & (e) DPSR(ours) \\
	\end{tabular}
	\caption{The visual comparison of different methods on ``\emph{chip}'' with scale factor 4 (first row), ``\emph{frog}'' with scale factor 3 (second row) and ``\emph{colour}'' with scale factor 2 (third row). The estimated blur kernel is shown on the upper-left corner of the LR image. Note that we assume
the noisy image ``frog'' has no blur.}
	\label{fig_visuafinal}\vspace{-0.3cm}
\end{figure*}

\vspace{-0.4cm}
\paragraph{Super-resolver prior with GAN.}
In the above experiments, SRResNet+ is trained by $\ell_1$ loss. It is natural to rise the question:  does the proposed method perform well if the super-resolver prior is trained with GAN loss?
Following the training strategy of~\cite{wang2018esrgan}, we trained a SRGAN+ model for scale factor 4 by a weighted combination of L1 loss, VGG loss and GAN loss.
For convenience, we refer to the proposed method with SRGAN+ as DPSRGAN.
In this special case, we compare DPSRGAN with IRCNN+RCAN and four GAN-related methods, including ESRGAN~\cite{wang2018esrgan}, IRCNN+ESRGAN and RCAN+DeblurGAN.

Fig.~\ref{fig:gan} shows the visual comparison of different methods.
It can be seen that directly super-resolving the blurry LR image by ESRGAN does not improve the visual quality.
By contrast, IRCNN+ESRGAN can deliver much better visual results as the distortion of blur is handled by IRCNN.
Meanwhile, it amplifies the perturbation error of IRCNN, resulting in unpleasant visual artifacts.
Although DeblurGAN is designed to handle motion blur, RCAN+DeblurGAN does not perform as well as expected.
In comparison, our DPSRGAN produces the most visually pleasant HR images with sharpness and naturalness.

\subsection{LR images with estimated kernel}\label{sec:blurry}
In this section, we focus on experiments on blurry LR images with estimated blur kernel.
Such experiments can help to evaluate the feasibility of the new degradation model, the practicability and kernel sensitivity of the proposed DPSR.
It is especially noteworthy that we do not know the HR ground-truth of the LR images.

Fig.~\ref{fig_visuafinal} shows the visual comparison with state-of-the-art SISR methods (\ie, RCAN~\cite{zhang2018image}, SRMD~\cite{zhang2017learning} and ZSSR~\cite{shocher2018zero})
on classical image ``\emph{chip}''~\cite{fattal2007image}, noisy image ``\emph{frog}''~\cite{lebrun2015noise} and blurry image ``\emph{colour}''~\cite{pan2014deblurring}.
For the ``\emph{chip}'' and ``\emph{colour}'', the blur kernel is estimated by~\cite{pan2014deblurring}.
For the noisy ``\emph{frog}'', we assume it has no blur and directly adopt our super-resolver prior to obtain the HR image.
Note that once the blur kernel is estimated, our DPSR can reconstruct HR images with different scale factors, whereas SRMD and ZSSR with Eqn.~\eqref{degradation:eq1} need to estimate a separate blur kernel for each scale factor.

From Fig.~\ref{fig_visuafinal}, we can observe that RCAN has very limited ability to deal with blur and noise because of its oversimplified bicubic degradation model. With a more general degradation model, SRMD and ZSSR yield better results than RCAN on ``\emph{chip}'' and ``\emph{frog}''. However, they cannot recover the latent HR image for the blurry image ``\emph{colour}'' which is blurred by a large complex kernel. In comparison, our DPSR gives rise to the most visually pleasant results. As a result, our new degradation model is a feasible assumption and DPSR is an attractive SISR method as it can handle a wide variety of degradations.

\section{Conclusion}

In this paper, we proposed a well-principled deep plug-and-play super-resolution method for LR image with arbitrary blur kernels.
We first design an alternative degradation model which can benefit existing blind deblurring methods for kernel estimation.
We then solve the corresponding energy function via half quadratic splitting algorithm so as to exploit the merits of plug-and-play framework.
It turns out we can explicitly handled the distortion of blur by a specialized modular. Such a distinctive merit actually enables the proposed method to deal with arbitrary blur kernels.
It also turns out that we can plug super-resolver prior rather than denoiser prior into the plug-and-play framework.
As such, we can fully exploit the advances of existing DNN-based SISR methods to design and train the super-resolver prior.
Extensive experimental results demonstrated the feasibility of the new degradation model and effectiveness of the proposed method for super-resolving LR images with arbitrary blur kernels.

\section{Acknowledgements}
This work is supported by HK RGC General Research Fund (PolyU 152216/18E) and National Natural Science Foundation of China (grant no.~61671182, 61872118, 61672446).

\clearpage

{\small
\bibliographystyle{ieee}
\bibliography{egbib}

\begin{thebibliography}{10}\itemsep=-1pt

\bibitem{afonso2010fast}
Manya~V Afonso, Jos{\'e}~M Bioucas-Dias, and M{\'a}rio~AT Figueiredo.
\newblock Fast image recovery using variable splitting and constrained
  optimization.
\newblock {\em IEEE transactions on image processing}, 19(9):2345--2356, 2010.

\bibitem{agustsson2017ntire}
Eirikur Agustsson and Radu Timofte.
\newblock Ntire 2017 challenge on single image super-resolution: Dataset and
  study.
\newblock In {\em IEEE Conference on Computer Vision and Pattern Recognition
  Workshops}, volume~3, pages 126--135, July 2017.

\bibitem{baker2002limits}
Simon Baker and Takeo Kanade.
\newblock Limits on super-resolution and how to break them.
\newblock {\em IEEE Transactions on Pattern Analysis and Machine Intelligence},
  24(9):1167--1183, 2002.

\bibitem{beck2009fast}
Amir Beck and Marc Teboulle.
\newblock A fast iterative shrinkage-thresholding algorithm for linear inverse
  problems.
\newblock {\em SIAM journal on imaging sciences}, 2(1):183--202, 2009.

\bibitem{bevilacqua2012low}
Marco Bevilacqua, Aline Roumy, Christine Guillemot, and Marie-Line~Alberi
  Morel.
\newblock Low-complexity single-image super-resolution based on nonnegative
  neighbor embedding.
\newblock In {\em British Machine Vision Conference}, 2012.

\bibitem{bigdeli2017deep}
Siavash~Arjomand Bigdeli, Meiguang Jin, Paolo Favaro, and Matthias Zwicker.
\newblock Deep mean-shift priors for image restoration.
\newblock In {\em Advances in Neural Information Processing Systems}, 2017.

\bibitem{boracchiFoiTIP12a}
G. Boracchi and A. Foi.
\newblock Modeling the performance of image restoration from motion blur.
\newblock {\em IEEE Transactions on Image Processing}, 21(8):3502 --3517, 2012.

\bibitem{boyd2011distributed}
Stephen Boyd, Neal Parikh, Eric Chu, Borja Peleato, and Jonathan Eckstein.
\newblock Distributed optimization and statistical learning via the alternating
  direction method of multipliers.
\newblock {\em Foundations and Trends in Machine Learning}, 3(1):1--122, 2011.

\bibitem{brifman2016turning}
Alon Brifman, Yaniv Romano, and Michael Elad.
\newblock Turning a denoiser into a super-resolver using plug and play priors.
\newblock In {\em IEEE International Conference on Image Processing}, pages
  1404--1408, 2016.

\bibitem{buzzard2018plug}
Gregery~T Buzzard, Stanley~H Chan, Suhas Sreehari, and Charles~A Bouman.
\newblock Plug-and-play unplugged: Optimization-free reconstruction using
  consensus equilibrium.
\newblock {\em SIAM Journal on Imaging Sciences}, 11(3):2001--2020, 2018.

\bibitem{chambolle2011first}
Antonin Chambolle and Thomas Pock.
\newblock A first-order primal-dual algorithm for convex problems with
  applications to imaging.
\newblock {\em Journal of Mathematical Imaging and Vision}, 40(1):120--145,
  2011.

\bibitem{chan2011single}
Stanley~H Chan and Truong~Q Nguyen.
\newblock Single image spatially variant out-of-focus blur removal.
\newblock In {\em IEEE International Conference on Image Processing}, pages
  677--680, 2011.

\bibitem{chan2016plug}
Stanley~H Chan, Xiran Wang, and Omar~A Elgendy.
\newblock {Plug-and-Play} {ADMM} for image restoration: Fixed-point convergence
  and applications.
\newblock {\em IEEE Transactions on Computational Imaging}, 3(1):84--98, 2017.

\bibitem{dabov2007image}
Kostadin Dabov, Alessandro Foi, Vladimir Katkovnik, and Karen Egiazarian.
\newblock Image denoising by sparse 3-{D} transform-domain collaborative
  filtering.
\newblock {\em IEEE Transactions on Image Processing}, 16(8):2080--2095, 2007.

\bibitem{danielyan2010image}
Aram Danielyan, Vladimir Katkovnik, and Karen Egiazarian.
\newblock Image deblurring by augmented lagrangian with {BM3D} frame prior.
\newblock In {\em Workshop on Information Theoretic Methods in Science and
  Engineering}, pages 16--18, 2010.

\bibitem{danielyan2012bm3d}
Aram Danielyan, Vladimir Katkovnik, and Karen Egiazarian.
\newblock {BM3D} frames and variational image deblurring.
\newblock {\em IEEE Transactions on Image Processing}, 21(4):1715--1728, 2012.

\bibitem{dong2014learning}
Chao Dong, Chen~Change Loy, Kaiming He, and Xiaoou Tang.
\newblock Learning a deep convolutional network for image super-resolution.
\newblock In {\em European Conference on Computer Vision}, pages 184--199,
  2014.

\bibitem{dong2016accelerating}
Chao Dong, Chen~Change Loy, and Xiaoou Tang.
\newblock Accelerating the super-resolution convolutional neural network.
\newblock In {\em European Conference on Computer Vision}, pages 391--407,
  2016.

\bibitem{dong2013nonlocally}
Weisheng Dong, Lei Zhang, Guangming Shi, and Xin Li.
\newblock Nonlocally centralized sparse representation for image restoration.
\newblock {\em IEEE Transactions on Image Processing}, 22(4):1620--1630, 2013.

\bibitem{efrat2013accurate}
Netalee Efrat, Daniel Glasner, Alexander Apartsin, Boaz Nadler, and Anat Levin.
\newblock Accurate blur models vs. image priors in single image
  super-resolution.
\newblock In {\em IEEE International Conference on Computer Vision}, pages
  2832--2839, 2013.

\bibitem{egiazarian2015single}
Karen Egiazarian and Vladimir Katkovnik.
\newblock Single image super-resolution via {BM3D} sparse coding.
\newblock In {\em European Signal Processing Conference}, pages 2849--2853,
  2015.

\bibitem{fattal2007image}
Raanan Fattal.
\newblock Image upsampling via imposed edge statistics.
\newblock In {\em ACM transactions on graphics}, volume~26, page~95, 2007.

\bibitem{gharbi2016deep}
Micha{\"e}l Gharbi, Gaurav Chaurasia, Sylvain Paris, and Fr{\'e}do Durand.
\newblock Deep joint demosaicking and denoising.
\newblock {\em ACM Transactions on Graphics}, 35(6):191, 2016.

\bibitem{goodfellow2014generative}
Ian Goodfellow, Jean Pouget-Abadie, Mehdi Mirza, Bing Xu, David Warde-Farley,
  Sherjil Ozair, Aaron Courville, and Yoshua Bengio.
\newblock Generative adversarial nets.
\newblock In {\em Advances in neural information processing systems}, pages
  2672--2680, 2014.

\bibitem{gu2018integrating}
Shuhang Gu, Radu Timofte, and Luc Van~Gool.
\newblock Integrating local and non-local denoiser priors for image
  restoration.
\newblock In {\em International Conference on Pattern Recognition}, 2018.

\bibitem{heide2014flexisp}
Felix Heide, Markus Steinberger, Yun-Ta Tsai, Mushfiqur Rouf, Dawid Pajak,
  Dikpal Reddy, Orazio Gallo, Jing Liu, Wolfgang Heidrich, Karen Egiazarian,
  et~al.
\newblock Flexisp: A flexible camera image processing framework.
\newblock {\em ACM Transactions on Graphics}, 33(6):231, 2014.

\bibitem{ioffe2015batch}
Sergey Ioffe and Christian Szegedy.
\newblock Batch normalization: Accelerating deep network training by reducing
  internal covariate shift.
\newblock In {\em International Conference on Machine Learning}, pages
  448--456, 2015.

\bibitem{jiang2018deep}
Junjun Jiang, Yi Yu, Jinhui Hu, Suhua Tang, and Jiayi Ma.
\newblock Deep {CNN} denoiser and multi-layer neighbor component embedding for
  face hallucination.
\newblock In {\em International Joint Conference on Artificial Intelligence},
  pages 771--778, 2018.

\bibitem{kim2015accurate}
Jiwon Kim, Jung~Kwon Lee, and Kyoung~Mu Lee.
\newblock Accurate image super-resolution using very deep convolutional
  networks.
\newblock In {\em IEEE Conference on Computer Vision and Pattern Recognition},
  pages 1646--1654, 2016.

\bibitem{kingma2014adam}
Diederik Kingma and Jimmy Ba.
\newblock Adam: A method for stochastic optimization.
\newblock In {\em International Conference for Learning Representations}, 2015.

\bibitem{kruse2017learning}
Jakob Kruse, Carsten Rother, and Uwe Schmidt.
\newblock Learning to push the limits of efficient fft-based image
  deconvolution.
\newblock In {\em IEEE International Conference on Computer Vision}, pages
  4586--4594, 2017.

\bibitem{kupyn2018deblurgan}
Orest Kupyn, Volodymyr Budzan, Mykola Mykhailych, Dmytro Mishkin, and Jiri
  Matas.
\newblock {DeblurGAN}: Blind motion deblurring using conditional adversarial
  networks.
\newblock In {\em IEEE International Conference on Computer Vision}, pages
  945--952, 2013.

\bibitem{lebrun2015noise}
Marc Lebrun, Miguel Colom, and Jean-Michel Morel.
\newblock The noise clinic: a blind image denoising algorithm.
\newblock {\em Image Processing On Line}, 5:1--54, 2015.

\bibitem{ledig2016photo}
Christian Ledig, Lucas Theis, Ferenc Husz{\'a}r, Jose Caballero, Andrew
  Cunningham, Alejandro Acosta, Andrew Aitken, Alykhan Tejani, Johannes Totz,
  Zehan Wang, et~al.
\newblock Photo-realistic single image super-resolution using a generative
  adversarial network.
\newblock In {\em IEEE Conference on Computer Vision and Pattern Recognition},
  pages 4681--4690, July 2017.

\bibitem{freeman2009understanding}
Anat Levin, Yair Weiss, Fredo Durand, and William~T Freeman.
\newblock Understanding and evaluating blind deconvolution algorithms.
\newblock pages 1964--1971, 2009.

\bibitem{lim2017enhanced}
Bee Lim, Sanghyun Son, Heewon Kim, Seungjun Nah, and Kyoung~Mu Lee.
\newblock Enhanced deep residual networks for single image super-resolution.
\newblock In {\em IEEE Conference on Computer Vision and Pattern Recognition
  Workshops}, pages 136--144, July 2017.

\bibitem{liu2018proximal}
Risheng Liu, Xin Fan, Shichao Cheng, Xiangyu Wang, and Zhongxuan Luo.
\newblock Proximal alternating direction network: A globally converged deep
  unrolling framework.
\newblock In {\em AAAI Conference on Artificial Intelligence}, pages
  1371--1378, 2018.

\bibitem{liu2019learning}
Risheng Liu, Long Ma, Yiyang Wang, and Lei Zhang.
\newblock Learning converged propagations with deep prior ensemble for image
  enhancement.
\newblock {\em IEEE Transactions on Image Processing}, 28(3):1528--1543, 2019.

\bibitem{MartinFTM01}
D. Martin, C. Fowlkes, D. Tal, and J. Malik.
\newblock A database of human segmented natural images and its application to
  evaluating segmentation algorithms and measuring ecological statistics.
\newblock In {\em IEEE International Conference on Computer Vision}, volume~2,
  pages 416--423, July 2001.

\bibitem{meinhardt2017learning}
Tim Meinhardt, Michael M{\"o}ller, Caner Hazirbas, and Daniel Cremers.
\newblock Learning proximal operators: Using denoising networks for
  regularizing inverse imaging problems.
\newblock In {\em IEEE International Conference on Computer Vision}, pages
  1781--1790, 2017.

\bibitem{michaeli2013nonparametric}
Tomer Michaeli and Michal Irani.
\newblock Nonparametric blind super-resolution.
\newblock In {\em IEEE International Conference on Computer Vision}, pages
  945--952, 2013.

\bibitem{ono2017primal}
Shunsuke Ono.
\newblock Primal-dual plug-and-play image restoration.
\newblock {\em IEEE Signal Processing Letters}, 24(8):1108--1112, 2017.

\bibitem{pan2014deblurring}
Jinshan Pan, Zhe Hu, Zhixun Su, and Ming-Hsuan Yang.
\newblock Deblurring text images via {L0}-regularized intensity and gradient
  prior.
\newblock In {\em IEEE Conference on Computer Vision and Pattern Recognition},
  pages 2901--2908, 2014.

\bibitem{plotz2018neural}
Tobias Pl{\"o}tz and Stefan Roth.
\newblock Neural nearest neighbors networks.
\newblock In {\em Advances in Neural Information Processing Systems}, pages
  1095--1106, 2018.

\bibitem{reehorst2018regularization}
Edward~T Reehorst and Philip Schniter.
\newblock Regularization by denoising: Clarifications and new interpretations.
\newblock {\em arXiv preprint arXiv:1806.02296}, 2018.

\bibitem{romano2017little}
Yaniv Romano, Michael Elad, and Peyman Milanfar.
\newblock The little engine that could: Regularization by denoising (red).
\newblock {\em SIAM Journal on Imaging Sciences}, 10(4):1804--1844, 2017.

\bibitem{rond2016poisson}
Arie Rond, Raja Giryes, and Michael Elad.
\newblock Poisson inverse problems by the plug-and-play scheme.
\newblock {\em Journal of Visual Communication and Image Representation},
  41:96--108, 2016.

\bibitem{roth2009fields}
Stefan Roth and Michael~J Black.
\newblock Fields of experts.
\newblock {\em International Journal of Computer Vision}, 82(2):205--229, 2009.

\bibitem{sajjadi2017enhancenet}
Mehdi~SM Sajjadi, Bernhard Sch{\"o}lkopf, and Michael Hirsch.
\newblock Enhancenet: Single image super-resolution through automated texture
  synthesis.
\newblock In {\em IEEE International Conference on Computer Vision}, pages
  4501--4510, 2017.

\bibitem{shi2016real}
Wenzhe Shi, Jose Caballero, Ferenc Husz{\'a}r, Johannes Totz, Andrew~P Aitken,
  Rob Bishop, Daniel Rueckert, and Zehan Wang.
\newblock Real-time single image and video super-resolution using an efficient
  sub-pixel convolutional neural network.
\newblock In {\em IEEE Conference on Computer Vision and Pattern Recognition},
  pages 1874--1883, 2016.

\bibitem{shocher2018zero}
Assaf Shocher, Nadav Cohen, and Michal Irani.
\newblock ``zero-shot'' super-resolution using deep internal learning.
\newblock In {\em IEEE International Conference on Computer Vision}, pages
  3118--3126, 2018.

\bibitem{simonyan2014very}
Karen Simonyan and Andrew Zisserman.
\newblock Very deep convolutional networks for large-scale image recognition.
\newblock {\em arXiv preprint arXiv:1409.1556}, 2014.

\bibitem{sroubek2008simultaneous}
F Sroubek, Gabriel Crist{\'o}bal, and J Flusser.
\newblock Simultaneous super-resolution and blind deconvolution.
\newblock In {\em Journal of Physics: Conference Series}, volume 124, page
  012048, 2008.

\bibitem{tao2018scale}
Xin Tao, Hongyun Gao, Xiaoyong Shen, Jue Wang, and Jiaya Jia.
\newblock Scale-recurrent network for deep image deblurring.
\newblock In {\em IEEE Conference on Computer Vision and Pattern Recognition},
  pages 8174--8182, 2018.

\bibitem{timofte2014a+}
Radu Timofte, Vincent De~Smet, and Luc Van~Gool.
\newblock A+: Adjusted anchored neighborhood regression for fast
  super-resolution.
\newblock In {\em Asian Conference on Computer Vision}, pages 111--126, 2014.

\bibitem{tirer2018image}
Tom Tirer and Raja Giryes.
\newblock Image restoration by iterative denoising and backward projections.
\newblock {\em IEEE Transactions on Image Processing}, 28(3):1220--1234, 2019.

\bibitem{venkatakrishnan2013plug}
Singanallur~V Venkatakrishnan, Charles~A Bouman, and Brendt Wohlberg.
\newblock Plug-and-play priors for model based reconstruction.
\newblock In {\em IEEE Global Conference on Signal and Information Processing},
  pages 945--948, 2013.

\bibitem{wang2018esrgan}
Xintao Wang, Ke Yu, Shixiang Wu, Jinjin Gu, Yihao Liu, Chao Dong, Yu Qiao, and
  Chen~Change Loy.
\newblock {ESRGAN}: Enhanced super-resolution generative adversarial networks.
\newblock In {\em The European Conference on Computer Vision Workshops}, 2018.

\bibitem{xu2014deep}
Li Xu, Jimmy~SJ Ren, Ce Liu, and Jiaya Jia.
\newblock Deep convolutional neural network for image deconvolution.
\newblock In {\em NIPS}, pages 1790--1798, 2014.

\bibitem{yang2014single}
Chih-Yuan Yang, Chao Ma, and Ming-Hsuan Yang.
\newblock Single-image super-resolution: A benchmark.
\newblock In {\em European Conference on Computer Vision}, pages 372--386,
  2014.

\bibitem{yang2008image}
Jianchao Yang, John Wright, Thomas Huang, and Yi Ma.
\newblock Image super-resolution as sparse representation of raw image patches.
\newblock In {\em IEEE Conference on Computer Vision and Pattern Recognition},
  pages 1--8, 2008.

\bibitem{zhang2017beyond}
Kai Zhang, Wangmeng Zuo, Yunjin Chen, Deyu Meng, and Lei Zhang.
\newblock Beyond a gaussian denoiser: Residual learning of deep {CNN} for image
  denoising.
\newblock {\em IEEE Transactions on Image Processing}, pages 3142--3155, 2017.

\bibitem{zhang2017learning}
Kai Zhang, Wangmeng Zuo, Shuhang Gu, and Lei Zhang.
\newblock Learning deep {CNN} denoiser prior for image restoration.
\newblock In {\em IEEE Conference on Computer Vision and Pattern Recognition},
  pages 3929--3938, July 2017.

\bibitem{zhang2018ffdnet}
Kai Zhang, Wangmeng Zuo, and Lei Zhang.
\newblock {FFDNet}: Toward a fast and flexible solution for {CNN} based image
  denoising.
\newblock {\em IEEE Transactions on Image Processing}, 27(9):4608--4622, 2018.

\bibitem{zhang2018learning}
Kai Zhang, Wangmeng Zuo, and Lei Zhang.
\newblock Learning a single convolutional super-resolution network for multiple
  degradations.
\newblock In {\em IEEE Conference on Computer Vision and Pattern Recognition},
  pages 3262--3271, 2018.

\bibitem{zhang2018gated}
Xinyi Zhang, Hang Dong, Zhe Hu, Wei-Sheng Lai, Fei Wang, and Ming-Hsuan Yang.
\newblock Gated fusion network for joint image deblurring and super-resolution.
\newblock In {\em British Machine Vision Conference}, 2018.

\bibitem{zhang2018image}
Yulun Zhang, Kunpeng Li, Kai Li, Lichen Wang, Bineng Zhong, and Yun Fu.
\newblock Image super-resolution using very deep residual channel attention
  networks.
\newblock In {\em Proceedings of the European Conference on Computer Vision},
  pages 286--301, 2018.

\bibitem{zhang2018residual}
Yulun Zhang, Yapeng Tian, Yu Kong, Bineng Zhong, and Yun Fu.
\newblock Residual dense network for image super-resolution.
\newblock In {\em IEEE Conference on Computer Vision and Pattern Recognition},
  2018.

\bibitem{zoran2011learning}
Daniel Zoran and Yair Weiss.
\newblock From learning models of natural image patches to whole image
  restoration.
\newblock In {\em IEEE International Conference on Computer Vision}, pages
  479--486, 2011.

\end{thebibliography}
}

\end{document}